\documentclass[runningheads]{llncs}

\usepackage[T1]{fontenc}
\usepackage{hyperref}
\usepackage{graphicx}
\usepackage{multirow}
\usepackage{enumitem}
\usepackage{authblk}
\usepackage{float}
\usepackage{xcolor}
\usepackage{subcaption}
\begin{document}
\title{Analyzing Emotions in Bangla Social Media Comments Using Machine Learning and LIME}

\author{Bidyarthi Paul\inst{1,2} \and
SM Musfiqur Rahman\inst{1} \and
Dipta Biswas\inst{1} \and
Md. Ziaul Hasan\inst{1} \and
Md. Zahid Hossain\inst{1}}

\authorrunning{B. Paul et al.}

\institute{Ahsanullah University of Science and Technology, Dhaka \and
\email{\{Corresponding author\} E-mail: \textcolor{blue}{bidyarthipaul01@gmail.com} \\ musfiqur0608@gmail.com, diptabiswas223@gmail.com, Ziaulhasanf@gmail.com, zahidd16@gmail.com}}

\maketitle              

\begin{abstract}
Research on understanding emotions in written language continues to expand, especially for understudied languages with distinctive regional expressions and cultural features, such as Bangla. This study examines emotion analysis using 22,698 social media comments from the EmoNoBa dataset. For language analysis, we employ machine learning models—Linear SVM, KNN, and Random Forest—with n-gram data from a TF-IDF vectorizer. We additionally investigated how PCA affects the reduction of dimensionality. Moreover, we utilized a BiLSTM model and AdaBoost to improve decision trees. To make our machine learning models easier to understand, we used LIME to explain the predictions of the AdaBoost classifier, which uses decision trees. With the goal of advancing sentiment analysis in languages with limited resources, our work examines various techniques to find efficient techniques for emotion identification in Bangla.

\end{abstract}
\section{Introduction}
The analysis of emotions in text has proven to be useful in the field of computational linguistics for solving a variety of problems, especially with English texts. (Yang et al., 2012)\cite{1} have been able to get emotions out of suicide notes, identification of offensive language by (Allouch et al., 2018)\cite{2}, and use sentiment analysis to help cancer patients (Sosea et al., 2020)\cite{3}. A lot of progress has been made in these areas thanks to projects like SemEval Affective Texts (Agirre et al., 2007)\cite{4}, SemEval Effects of Tweets (Mohammad et al., 2018)\cite{5}, and GoEmotion (Demszky et al., 2020)\cite{6} that do a lot of work on difficult, multi-label emotion detection tasks. These efforts have cleared the path for significant advancements in understanding and interpreting human emotions through text.

Despite the progress in languages like English, Bangla, the sixth most spoken language worldwide\footnote{\url{https://en.wikipedia.org/wiki/List_of_ languages_by_total_numbers }} and the primary language in Bangladesh, has seen limited exploration in this domain. As Bangladesh emerges as a middle-income nation with expanding technology access even in rural areas (Basunia, 2022)\cite{7}, the importance of analyzing Bangla text for emotional content becomes increasingly relevant. Such analyses can significantly impact social welfare and business sectors by providing insights into public sentiment and emotional responses.

This research paper explores the analysis of emotions within Bangla text, utilizing a dataset of 22,698 public comments from social media (Islam et al., 2022)\cite{8}. These comments cover a diverse range of topics, including personal issues, politics, and health. The study examines various machine learning models, such as Linear SVM, KNN, and Random Forest, applying n-gram and TF-IDF vectorization to capture the linguistic nuances across unigrams, bigrams, and trigrams. Additionally, the role of PCA in dimensionality reduction is investigated, comparing results with and without its application. The study extends to decision trees enhanced with AdaBoost and explores deep learning through a BiLSTM model to gain a deeper understanding of semantic relationships in Bangla text. Local Interpretable Model-agnostic Explanations (LIME) is used to make the machine learning models more interpretable by explaining the predictions made by the AdaBoost classifier and decision trees. This comprehensive approach aims to establish a benchmark in emotion detection for Bangla, contributing to the broader field of sentiment analysis in under-researched languages. Some notable objectives of our study are given below:

\begin{enumerate}
    \item To evaluate the effectiveness of different machine learning models—including Linear SVM, KNN, and Random Forest—in detecting emotions in Bangla text to identify the most suitable algorithms for this task.

    \item To explore the role of Principal Component Analysis (PCA) in improving the performance of emotion detection models through dimensionality reduction.

    \item To enhance the accuracy of emotion detection by applying advanced techniques such as AdaBoost and BiLSTM, demonstrating their effectiveness in the context of Bangla text.

    \item To improve the interpretability of emotion detection models by utilizing Local Interpretable Model-agnostic Explanations (LIME), making the models' predictions more transparent.

    \item To establish a benchmark for emotion detection in Bangla, contributing to the broader field of Natural Language Processing (NLP) for low-resource languages.
    
\end{enumerate}

\section{Literature Review}
In the work of identifying specific emotions in Bangla text, Rahman et al. (2019)~\cite{9} conducted a thorough study, going beyond the simple classification of sentiments as either positive or negative. The authors carefully selected and provided detailed explanations for comments from Bangla Facebook groups that were involved in conversations about social and political topics. The process of examining emotions such as happiness, sadness, disgust, surprise, fear, and anger occurred. The study effectively employed machine learning techniques, specifically Support Vector Machines (SVM), to attain a significant accuracy rate of 53 percent in categorizing emotions. This study offers valuable insights into the nuanced comprehension of emotions in Bangla text.\\
A crucial task for sentiment analysis, identifying emotions in written text, was tackled by Yang et al. (2012)\cite{1}. Their work revolved around competing in an event devoted to identifying the emotions found in medical notes—especially those related to suicide. The system in question combined several language models to carry out tasks like sentiment analysis of sentences, word recognition, and machine learning-based emotion classification. By combining these various techniques, the study aimed to precisely identify and categorize 15 different emotions in suicide notes.\\
In a distinct domain, Allouch et al. (2018)\cite{2} concentrated on developing an agent-based system tailored to assist children with Autism Spectrum Disorder (ASD) in navigating their social interactions. The research specifically addressed situations wherein ASD children unintentionally uttered insulting sentences. The authors compiled a dataset comprising insulting and non-insulting sentences with input from parents of ASD children. Machine learning methods, including SVM, Multi-Layer Neural Network, and Tree Bagger, were employed to predict insulting sentences with precision and recall exceeding 75 percent. The study underscores the significance of automated agents in furnishing pertinent feedback and support for ASD children in social communication, especially in scenarios involving potential insults.\\
A dataset drawn from prominent newspapers and news websites is used by Agirre et al. (2007) \cite{4}. The work focuses on the difficult categorization of valence and emotions in these headlines, examining the complex relationship between lexical semantics and emotions. Five different models are included in the study, each of which adds to the rich field of automatic emotion recognition. These models are: UPAR7, which uses a linguistic approach based on rules; SICS, which uses a word-space model and seed words; CLaC, which presents a supervised Naïve Bayes classifier (CLaC-NB) and an unsupervised approach based on knowledge; UA, which uses statistics from web search engines; and SWAT, a supervised system that uses a unigram model with synonym expansion. This collection of models provides insights into different approaches and how well they perform in the given task, reflecting a thorough investigation of emotion recognition in affective text.\\
A new dataset for fine-grained emotion analysis in Bangla is presented by Islam et al. (2022)\cite{8}. Based on the Junto Emotion Wheel, EmoNoBa labels 22,698 Bangla social media comments from 12 domains for six fine-grained emotion categories. Working with low-resourced Bangla language and emotion dataset limitations are highlighted. The dataset is carefully prepared to preserve linguistic richness and challenge classification models. Language features, recurrent neural networks, and pre-trained language models are tested, and random baselines outperform them. The authors share the dataset and models for research. A lot of data was collected, labels were added, and linguistic features were analyzed in great detail. This gave researchers studying Bengali language emotions useful information. It also showed how important transformer-based models are for better understanding of context.

\section{Dataset}

\subsection{Dataset Collection}
\begin{figure}[h]
\centering
    \includegraphics[width=0.5\textwidth]{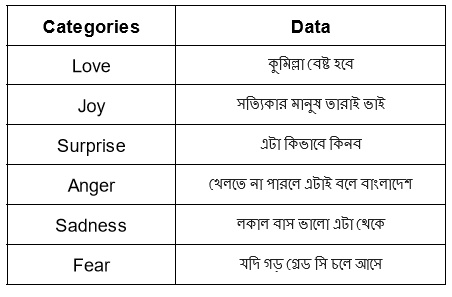}
    \caption{Sample Dataset}
    \label{fig:Cat}
\end{figure}

The EmoNoBa dataset is a comprehensive collection of 22,698 public comments in Bangla, sourced from Kaggle\footnote{\url{https://www.kaggle.com/datasets/saifsust/emonoba}}. This dataset was meticulously compiled to support research in emotion analysis within the Bangla-speaking community. It encompasses comments spanning 12 different domains (Personal, Politics, Sports, etc.), labeled for 6 different emotion categories \textbf{(love, joy, surprise, anger, sadness, fear)}, which has been illustrated in Figure~\ref{fig:Cat}.

\subsection{Dataset Objective:}
For six basic emotion categories, the aim is to figure out all the feelings expressed in a piece of text.

\subsection{Dataset Statistics and Analysis:}
Our dataset consists of 22,698 entries. On average, each entry is about 1.36 ± 0.82 sentences long. The typical sentence length is roughly 11.70 ± 10.70 words. The majority of the data, 77.28\% of the instances are sourced from YouTube, with the remaining data collected from Facebook and Twitter, covering 12 of the most popular topics from Prothom Alo\footnote{\url{https://www.prothomalo.com/}}. Additionally, 15.3\% of the entries express more than one emotion. 

\begin{figure}[h]
    \centering
    \includegraphics[width=0.8\textwidth]{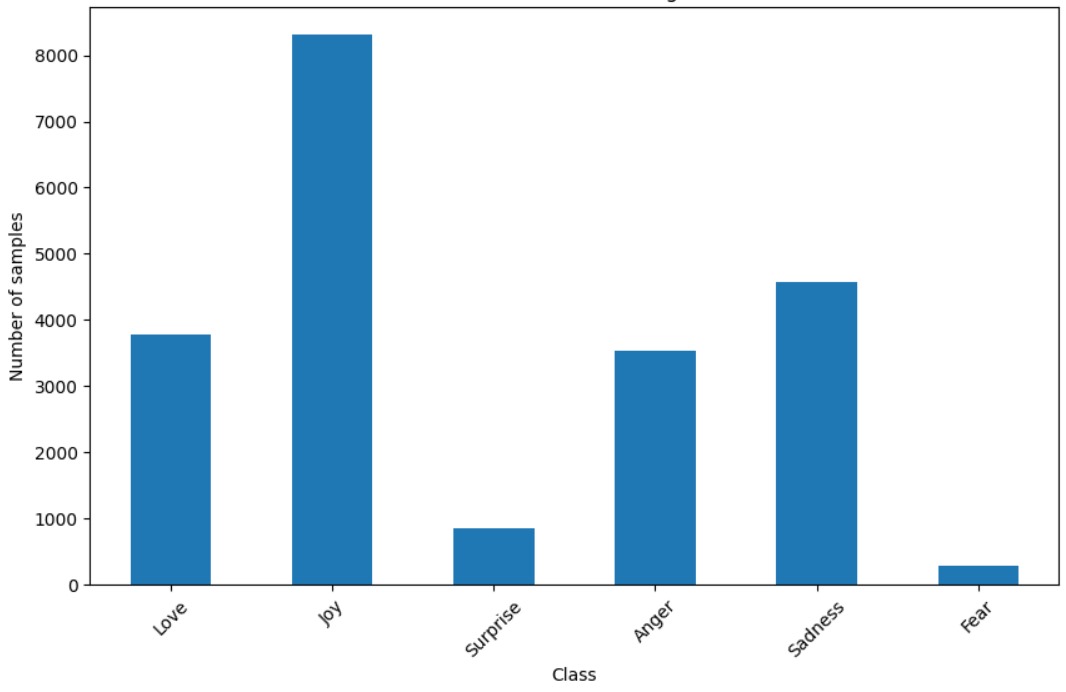}
    \caption{Number of samples in each class}
    \label{fig:bar}
\end{figure}
Figure~\ref{fig:bar} represents a bar chart illustrating the distribution of samples across different emotional categories within the dataset. The graph depicts six emotions: Love, Joy, Surprise, Anger, Sadness, and Fear. From the visualization, we observe that `Joy' has the highest number of instances, significantly more than the other emotions, followed by `Sadness' which also has a large number of samples. `Love' and `Anger' are represented in a moderate number of instances. In contrast, `Surprise' and `Fear' are the least represented emotions in the dataset, with `Fear' having the fewest instances. This distribution highlights the variability of emotional expressions in the dataset and indicates which emotions are more frequently expressed in the collected samples.

\begin{figure}[h]
    \centering
    \includegraphics[width=0.7\textwidth]{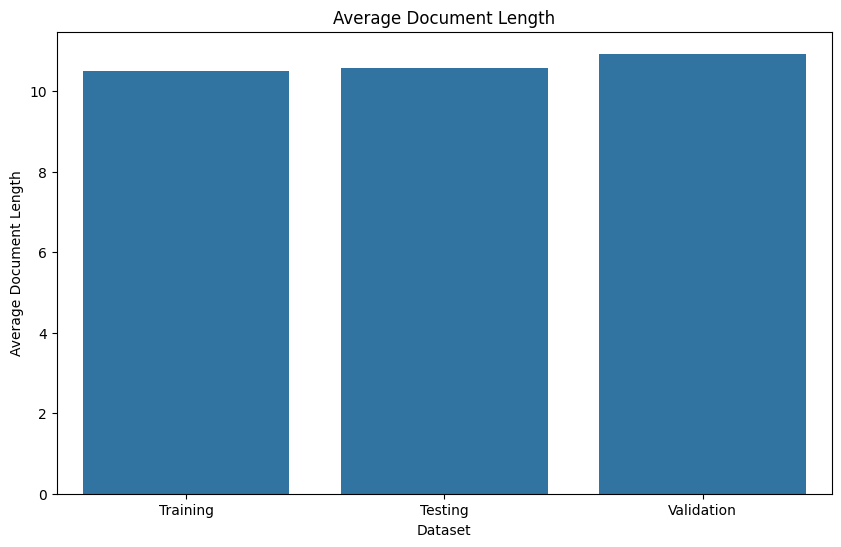}
    \caption{Average Length of the Documents}
    \label{fig:doc}
\end{figure}

\subsection{Train Test Split}
We conducted a stratified split for Machine Learning Models where 80\% was for training set, 15\% was for testing set and rest 5\% for the validation set.

Figure~\ref{fig:doc} illustrates the average length of the Bangla Text in each subset (Training, Test, Validation) of the dataset, which appears to be consistent across the three subsets, indicating that the documents are, on average, of similar length whether they are part of the training, testing, or validation set. 

\subsection{Word Cloud}

Figure~\ref{fig:word} represents a word cloud which is a visual representation of the most frequently occurring words within the dataset used for our research. Each word's size is proportional to its frequency: the larger the word, the more often it appears in the dataset. By visually representing word frequency, it helps in identifying key terms and common themes present in the text data. This is particularly useful in emotion analysis as it highlights the dominant words that may be associated with specific emotions, offering insights into the linguistic patterns prevalent in the dataset.

\subsection{t-SNE (t-distributed Stochastic Neighbor Embedding)}
\begin{figure}[h]
    \centering
    \begin{minipage}[b]{0.45\textwidth}
        \centering
        \includegraphics[width=\textwidth]{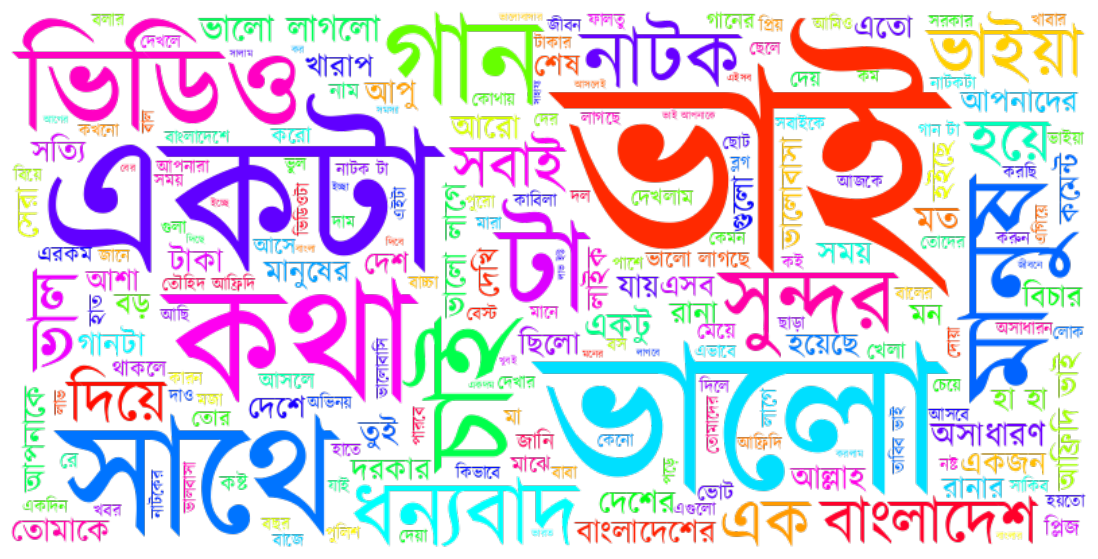}
        \caption{Word Cloud}
        \label{fig:word}
    \end{minipage}
    \hfill
    \begin{minipage}[b]{0.45\textwidth}
        \centering
        \includegraphics[width=\textwidth]{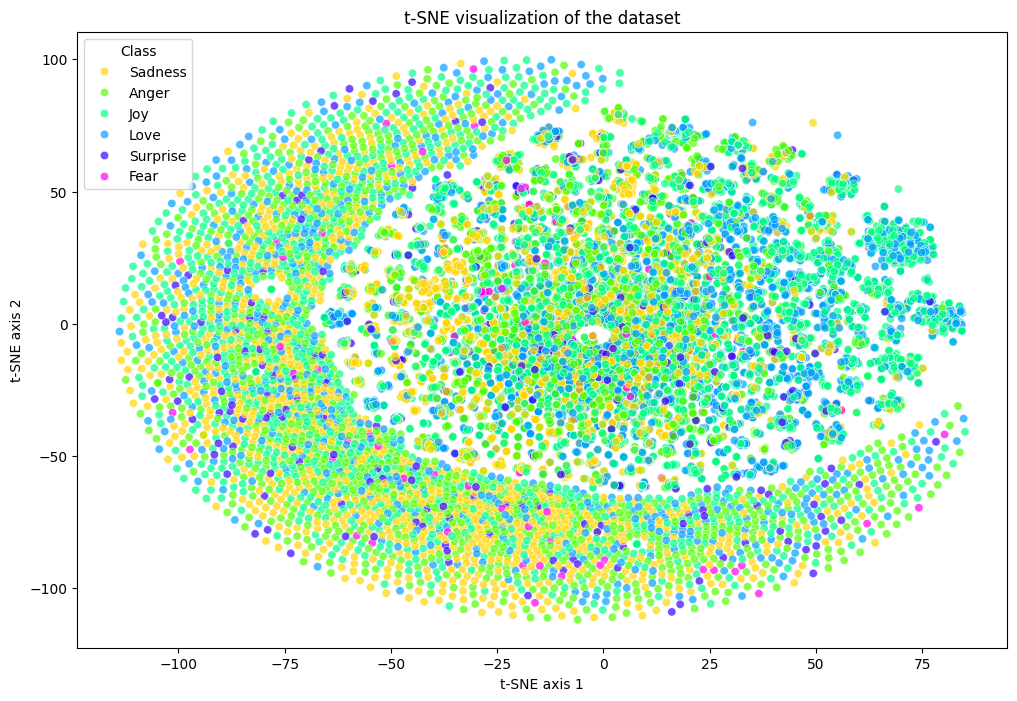}
        \caption{t-SNE}
        \label{fig:sne}
    \end{minipage}
\end{figure}

t-SNE is a dimensionality reduction technique that allows the visualization of high-dimensional data in a lower-dimensional space, typically two or three dimensions. It is particularly useful for visualizing the structure of data by preserving the local relationships between data points, making it easier to identify patterns or clusters that might not be apparent in higher dimensions. Figure~\ref{fig:sne} shows a t-SNE visualization of our dataset, where we've transformed the text data into numerical values using TF-IDF (Term Frequency-Inverse Document Frequency) before applying t-SNE. This visualization helps in understanding how well the emotions are separated or clustered based on the features extracted from the text data.

\section{Methodology}
\begin{figure}[h]
    \centering
    \includegraphics[width=1.0\textwidth]{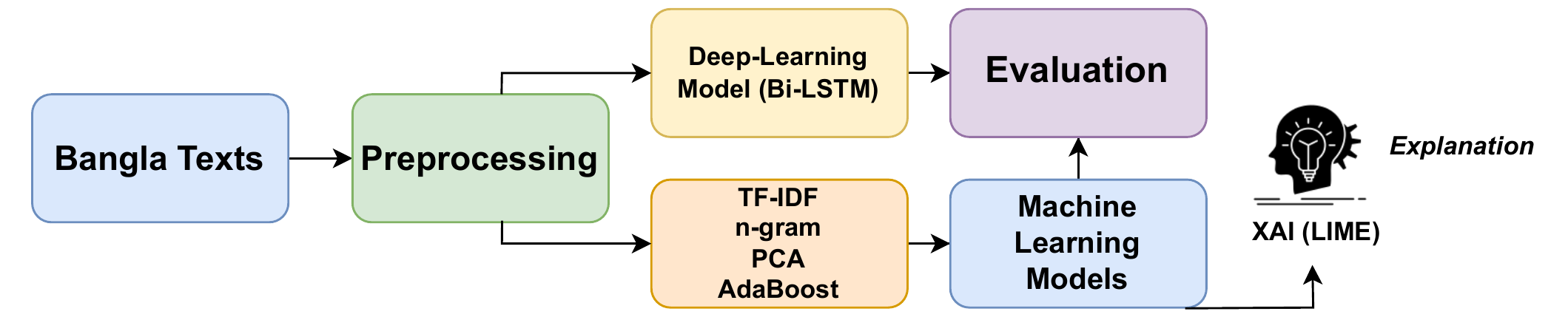}
    \caption{Methodology}
    \label{fig:flow}
\end{figure}

The methodology followed in our study is shown in Figure~\ref{fig:flow}. Here are some of the following steps that have been implemented in the study.

\subsection{Dataset Preprocessing}

For preprocessing our dataset, we focused on cleaning the text to ensure consistency and accuracy for analysis. This involved removing all punctuation marks and emojis, as these elements could interfere with our emotion analysis algorithms. We also addressed the issue of multiple spaces within the comments, reducing them to single spaces to maintain uniformity in the text data. These steps helped to streamline the dataset, making it more straightforward for the computational models to process.

\subsection{Classification}
\subsubsection{\textbf{Machine Learning Approach:}}

\begin{figure}[h]
    \centering
    \includegraphics[width=0.9\textwidth]{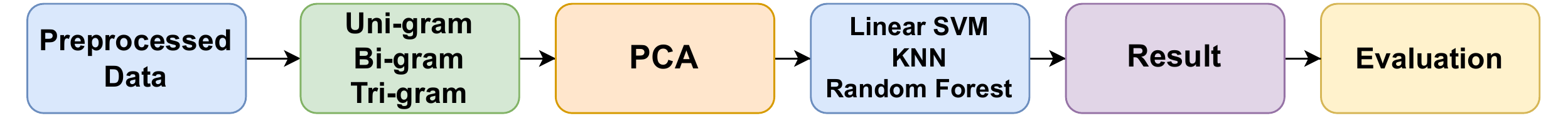}
    \caption{Flow Chart of Machine Learning Approach (PCA and n-gram)}
    \label{fig:ml}
\end{figure}

Figure~\ref{fig:ml} and~\ref{fig:ada} show an overview of our machine learning approach. This study employs four well-established models for emotion classification in text: Linear Support Vector Machine (SVM), k-Nearest Neighbors (kNN), Random Forest Classifier, and Decision Tree. The selection of these models is grounded in their proven strengths in text classification tasks. SVM is particularly effective in handling high-dimensional spaces, such as those created by TF-IDF vectors, and its regularization capability helps control overfitting in scenarios with large feature sets~\cite{meth1}. kNN is chosen for its simplicity and effectiveness in classifying similar instances in close proximity, making it suitable for emotion classification where text samples exhibit proximity-based similarities~\cite{meth2}. Random Forest is included for its robustness and ability to manage imbalanced datasets, which is often the case in emotion classification where certain emotions may be underrepresented~\cite{meth3}. Lastly, Decision Trees were selected for their interpretability and versatility, making them ideal for understanding the underlying decision processes in emotion classification~\cite{meth4}. The approaches described in i, ii, and iii were integrated into our machine learning models to evaluate their impact on classification performance and model efficiency.

\begin{enumerate}[label=\roman*]
    \item \textbf{n-gram}: To assess the influence of text structure on model performance, we applied the aforementioned n-gram techniques. This considers single words, pairs, and triplets of consecutive words as unique features. This method allows for a nuanced analysis of the linguistic patterns that may denote emotional states.
    
    \item \textbf{PCA}: With an aim to explore the potential benefits of dimensionality reduction on our models' predictive accuracy, implementation of Principal Component Analysis (PCA) was done. PCA simplifies the complexity of high-dimensional data while preserving as much variability as possible, facilitating a more streamlined and potentially more insightful analysis.
    
    \item \textbf{AdaBoost}: The study further explored the use of AdaBoost with a Decision Tree model to evaluate its impact on classification performance. AdaBoost enhances the model by focusing on instances that are harder to classify, combining multiple weak learners to create a stronger overall classifier. This approach helps to improve accuracy and reduce errors, making it particularly effective in refining the model’s ability to classify emotions in text. By adjusting the model’s focus on misclassified instances, AdaBoost boosts the overall performance and generalization of the Decision Tree. Figure~\ref{fig:ada}, illustrates the workflow of AdaBoost in our methodology for the Decision Tree model.

    \begin{figure}[h]
    \centering
    \includegraphics[width=0.9\textwidth]{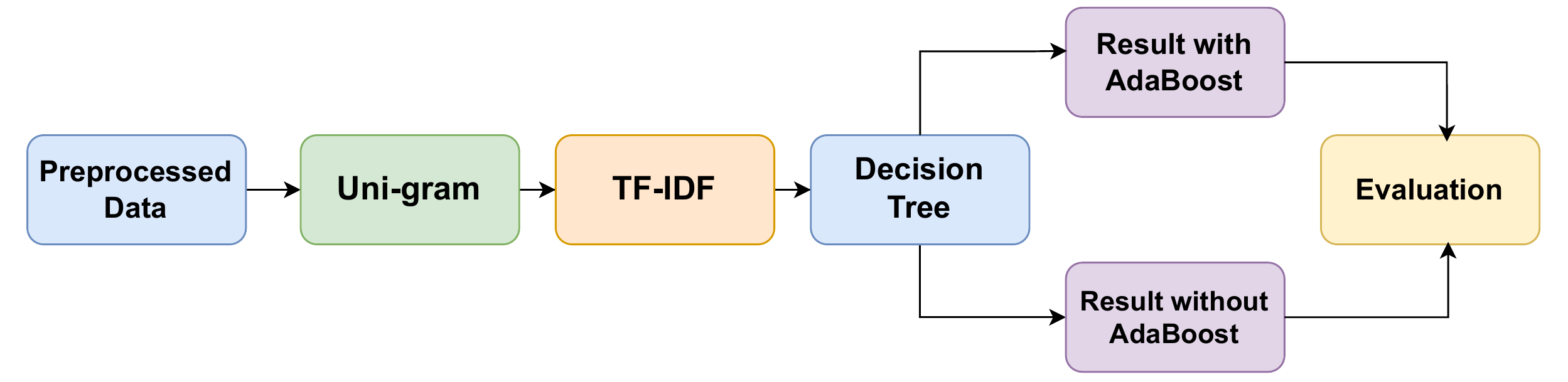}
    \caption{Flow Chart of Machine Learning Approach (AdaBoost and n-gram)}
    \label{fig:ada}
    \end{figure}

    \item \textbf{Local Interpretable Model-agnostic Explanations (LIME)} can be used for enhancing the interpretability of complex machine learning models used for emotion classification. Models like AdaBoost and deep learning can achieve high accuracy but often act as "black boxes," making their decision-making process unclear. LIME helps by providing explanations that show which words or features were most influential in the model's decisions.

In our study, LIME ensures that predictions are not only accurate but also understandable. This is crucial for interpreting emotional content in text, as it allows us to see how the model makes its classifications. By using LIME, we verify that the model is focusing on relevant features, increasing the trustworthiness and transparency of the results. This interpretability is essential for validating the model's effectiveness in capturing subtle emotional cues in Bangla.
    
\end{enumerate}

\subsubsection{\textbf{Deep Learning Approach:}}

\begin{figure}[h]
    \centering
    \includegraphics[width=0.8\textwidth]{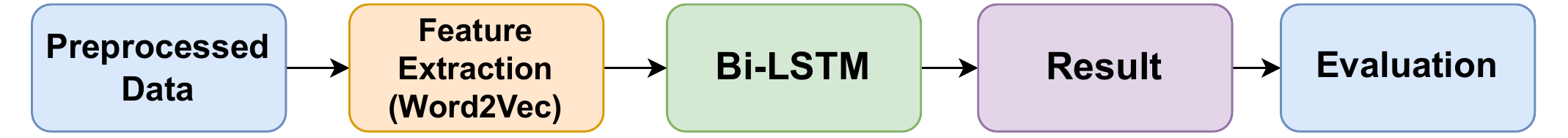}
    \caption{Flow Chart of Deep Learning Approach}
    \label{fig:dl}
\end{figure}

For our deep learning approach, we opted for a Bidirectional Long Short-Term Memory (BiLSTM) network (Mojumder, Pritom, et al.)\cite{12}. The BiLSTM model, renowned for its proficiency in capturing sequential information from text in both forward and backward directions (Hochreiter et al.)\cite{10}, served as the backbone of our architecture. To extract features from the text, we employed Word2Vec embeddings, which transformed words into dense vector representations, preserving their semantic relationships. These embeddings were then fed into the BiLSTM model to enhance its ability to learn contextual information from the input data. The entire workflow is illustrated in Figure~\ref{fig:dl}.

\subsection{Evaluation}
To evaluate the performance of our classification model, we deployed a comprehensive set of metrics, including accuracy, precision, recall, and the F1 score. These metrics were calculated in three distinct forms to provide a distinct view of the model's performance: macro, micro, and weighted averages. Table~\ref{tab:pca} Shows our evaluation metrics.

\begin{table}[h]
\centering
\caption{Evaluation Metrics}
\label{tab:pca}
\begin{tabular}{|c|c|c|c|p{5cm}}
\hline
\textbf{Metrics} & \textbf{Micro} & \textbf{Macro} & \textbf{Weighted}\\
\hline
Accuracy & N/A & N/A & N/A  \\
\hline
Precision & Yes & Yes &Yes \\
\hline
Recall & Yes & Yes & Yes \\
\hline 
F1-score & Yes & Yes & Yes \\
\hline

\end{tabular}
\end{table}

\subsection{Experimental setup}
Our work was carried out in a Google Colab Notebook using Python 3.10.12, PyTorch 2.0.1, with a Tesla T4 GPU (15 GB), 12.5 GB of RAM, and 64 GB of disk space.

\section{Result analysis}

\subsubsection{Machine Learning Based Results:}

\begin{table}[H]
\caption{Result overview of Linear SVM, KNN and random Forest}
\label{tab:my-tableres}
\resizebox{\columnwidth}{!}{%
\begin{tabular}{|c|cccccc|}
\hline
\multirow{2}{*}{\textbf{Model}} & \multicolumn{6}{c|}{\textbf{F1-Score}}                                                                                                                                                                                                         \\ \cline{2-7} 
                                & \multicolumn{1}{c|}{\textbf{Uni-gram (PCA)}} & \multicolumn{1}{c|}{\textbf{Uni-gram}} & \multicolumn{1}{c|}{\textbf{Bi-gram (PCA)}} & \multicolumn{1}{c|}{\textbf{Bi-gram}} & \multicolumn{1}{c|}{\textbf{Tri-gram (PCA)}} & \textbf{Tri-gram} \\ \hline
Linear SVM                      & \multicolumn{1}{c|}{0.56}                    & \multicolumn{1}{c|}{\textbf{0.63}}     & \multicolumn{1}{c|}{0.50}                   & \multicolumn{1}{c|}{0.57}             & \multicolumn{1}{c|}{0.46}                    & 0.49              \\ \hline
KNN                             & \multicolumn{1}{c|}{0.54}                    & \multicolumn{1}{c|}{\textbf{0.57}}     & \multicolumn{1}{c|}{0.52}                   & \multicolumn{1}{c|}{0.55}             & \multicolumn{1}{c|}{0.46}                    & 0.46              \\ \hline
Random Forest                   & \multicolumn{1}{c|}{0.55}                    & \multicolumn{1}{c|}{\textbf{0.57}}     & \multicolumn{1}{c|}{0.52}                   & \multicolumn{1}{c|}{0.52}             & \multicolumn{1}{c|}{0.45}                    & 0.44              \\ \hline
\end{tabular}%
}
\end{table}

Table~\ref{tab:my-tableres} indicates that the Linear SVM model consistently outperforms both KNN and Random Forest across various n-gram configurations, achieving its highest F1-score of 0.63 with the Uni-gram setup. This suggests that simpler text structures (like Uni-grams) are more effective for emotion classification in this context. Both KNN and Random Forest show similar patterns, with their best performances also occurring with the Uni-gram configuration, but their overall F1-scores are slightly lower compared to SVM, particularly when using bi-grams and tri-grams, which seem to introduce complexity without significant performance gains. Additionaly, applying PCA generally led to a reduction in model performance across all classifiers. For instance, the Linear SVM model's F1-score dropped from 0.63 to 0.56 when PCA was applied to Uni-grams. Similar declines were observed with KNN and Random Forest, indicating that while PCA helps in reducing dimensionality, it might also remove valuable information in the text features, leading to less effective emotion classification. This suggests that the original feature space without reducing the dimensionality better captures the details needed for distinguishing emotional states in text.

\begin{table}[h]
\centering
\caption{Decision Tree based results}
\label{tab:tab5}
\begin{tabular}{|c|c|p{6cm}}
\hline
\textbf{AdaBoost} &  \textbf{Macro average F1-score}\\
\hline
With AdaBoost & 0.7860 \\
\hline
Without AdaBoost  & 0.7799\\
\hline

\end{tabular}
\end{table}

Table~\ref{tab:tab5} shows that incorporating AdaBoost with the Decision Tree model results in a slight improvement in the macro-average F1-score from 0.7799 to 0.7860. This indicates that while the base Decision Tree model performs well, AdaBoost helps by slightly enhancing its ability to handle more challenging classifications. However, the improvement is modest, suggesting that the Decision Tree was already capturing most of the data patterns effectively.

\begin{figure}[h]
    \centering
    \begin{minipage}{0.8\textwidth}
        \centering
        \includegraphics[width=\textwidth]{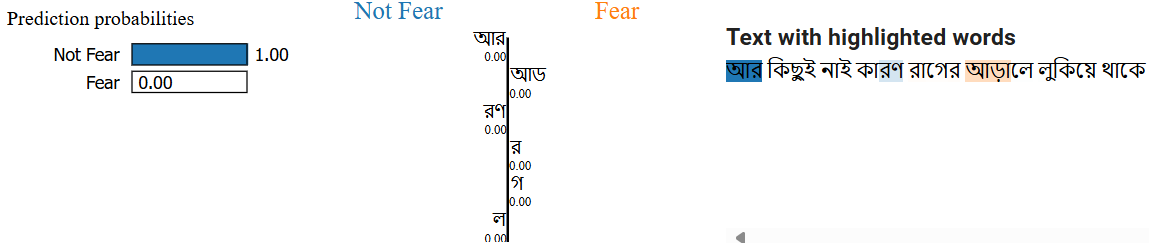} 
  
    \end{minipage}\hfill
    \begin{minipage}{0.8\textwidth}
        \centering
        \includegraphics[width=\textwidth]{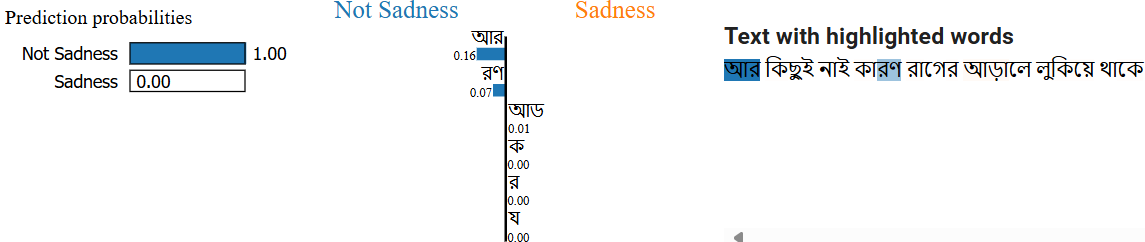} 
     
    \end{minipage}\hfill
    \begin{minipage}{0.8\textwidth}
        \centering
        \includegraphics[width=\textwidth]{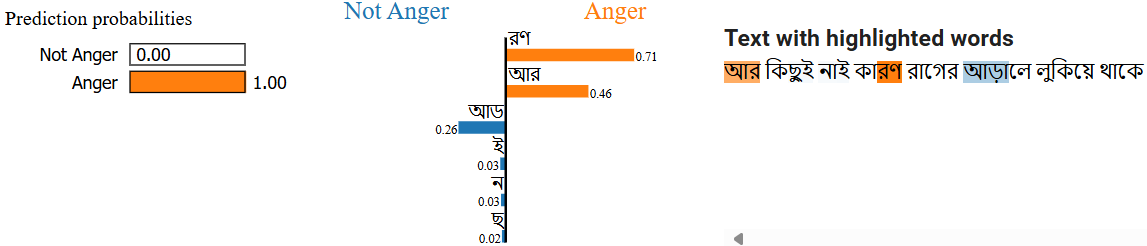} 
    
    \end{minipage}\hfill
    \begin{minipage}{0.8\textwidth}
        \centering
        \includegraphics[width=\textwidth]{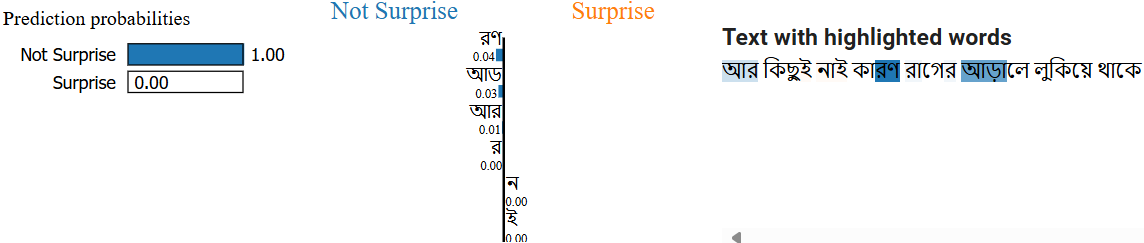} 
    \end{minipage}\hfill
    \begin{minipage}{0.8\textwidth}
        \centering
        \includegraphics[width=\textwidth]{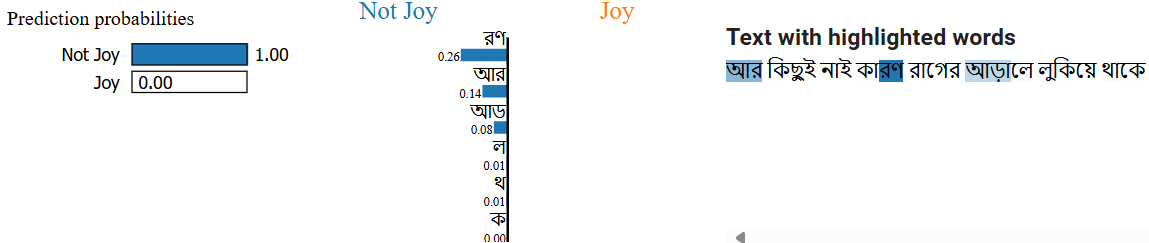} 
    \end{minipage}\hfill
    \begin{minipage}{0.8\textwidth}
        \centering
        \includegraphics[width=\textwidth]{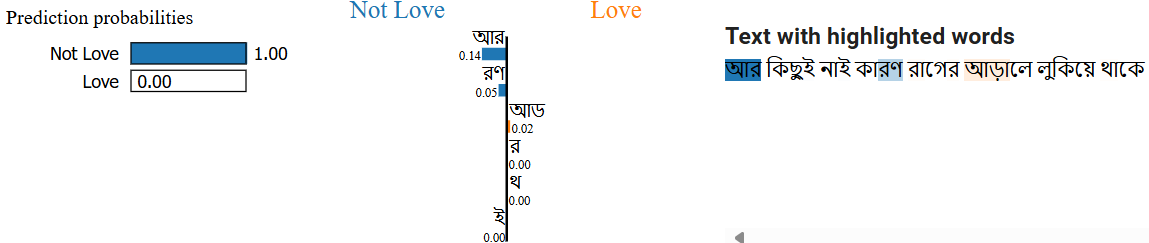} 
    \end{minipage}
    \caption{Sample of LIME Explanations for all emotions}
    \label{fig:sidebyside}
\end{figure}

\begin{table}[h]
\centering
\caption{Deep Learning based results}
\label{tab:tab6}
\begin{tabular}{|c|c|p{6cm}}
\hline
\textbf{Model} &  \textbf{F1-score}\\
\hline
Bi-LSTM & 0.3869 \\
\hline

\end{tabular}
\end{table}

\subsubsection{Deep Learning Based Results:}

Here, Table~\ref{tab:tab6} represents the results for the Bi-LSTM deep learning model. This model achieved an F1-score of 0.3869. The score reflects the average effectiveness of the model across different emotional categories within the dataset. Despite being a sophisticated deep learning approach, the F1-score indicates there is significant room for improvement in the model's performance.

\subsection{Model Performance Summary}

\begin{table}[h]
\centering
\caption{Summary of Best Model Performances}
\label{tab:tab7}
\begin{tabular}{|c|c|c|c|c|p{3cm}}
\hline
\textbf{Model} & \textbf{n-Gram} & \textbf{PCA} & \textbf{AdaBoost} & \textbf{F1-score}\\
\hline
SVM & Uni-gram & No & No & 0.63 \\
\hline
KNN & Uni-gram & No & No & 0.57 \\
\hline
Random Forest & Uni-gram & No & No & 0.57 \\
\hline 
\textbf{Decision Tree} & Uni-gram & No & Yes & \textbf{0.7860}\\
\hline
Bi-LSTM & No & No & No & 0.3869\\
\hline
\end{tabular}
\end{table}

The analysis of Table~\ref{tab:tab7} shows that the Linear SVM model achieved the highest F1-score of 0.63 using Uni-grams without PCA or AdaBoost, outperforming KNN and Random Forest, both of which also performed best with Uni-grams at 0.57. The Decision Tree, when combined with AdaBoost, recorded the highest overall F1-score of 0.7860, highlighting the effectiveness of boosting in improving model accuracy. In contrast, the Bi-LSTM model, with an F1-score of 0.3869, underperformed compared to the traditional models. It suggests that simpler, traditional methods were more effective for this specific text classification task.

\begin{figure}[h]
    \centering
    \begin{minipage}[b]{0.4\textwidth}
        \centering
        \includegraphics[width=\textwidth]{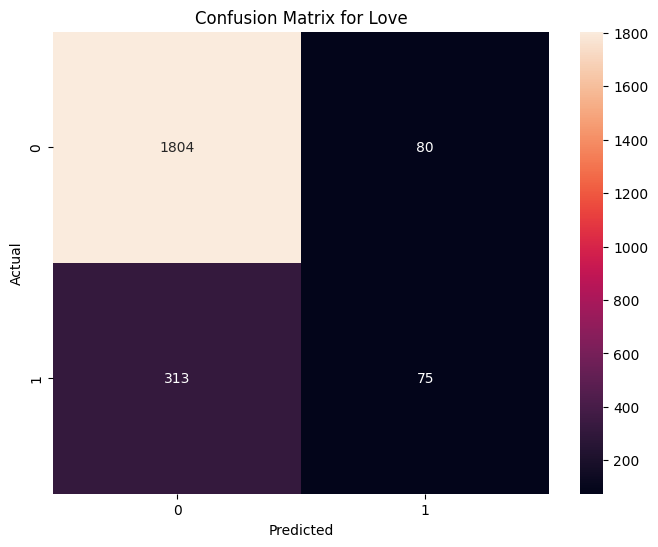}
        \caption{Confusion Matrix of Love}
        \label{fig:10}
    \end{minipage}
    \hfill
    \begin{minipage}[b]{0.4\textwidth}
        \centering
        \includegraphics[width=\textwidth]{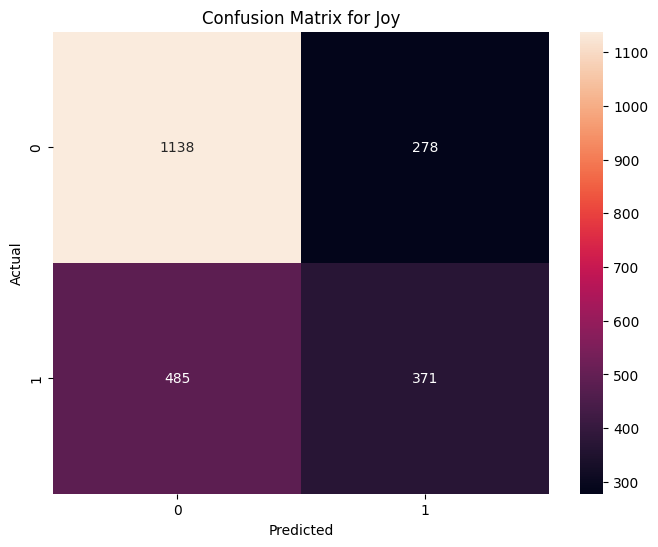}
        \caption{Confusion Matrix of Surprise}
        \label{fig:11}
    \end{minipage}
\end{figure}

Figure~\ref{fig:10} and ~\ref{fig:11} show us a sample of confusion matrix of 2 labels for our best performing model (Decision Tree).

\subsubsection{Lime Explanation:}From Table~\ref{tab:tab7}, it can be seen that Decision Tree with AdaBoost performs the best. LIME has been implemented in that model, to analyze how it accurately predicts emotion classifications. This allowed us to identify the key features that influenced the model's decisions, enhancing the interpretability of the predictions. From the Figure~\ref{fig:sidebyside}, we can observe that, the model confidently predicts "Not Fear," "Not Sadness," "Not Surprise," "Not Joy," and "Not Love" for the given text but classifies it as containing "Anger" with high certainty. The use of LIME helps us understand which specific words are contributing to these predictions.

% \usepackage{graphicx}
\begin{table}[]
\caption{An overview of some research articles that worked on the EmoNoBa dataset, including their corresponding benchmark models
and assessments}
\label{tab:my-table12}
\resizebox{\columnwidth}{!}{%
\begin{tabular}{|c|c|c|}
\hline
\textbf{Study}                                                 & \textbf{Model}                                                                               & \textbf{F1-Score} \\ \hline
\begin{tabular}[c]{@{}c@{}}Islam et al.~\cite{8} \\ (2022)\end{tabular} & \begin{tabular}[c]{@{}c@{}}Lexical Feature\\ (W1 + W2 + W3 + W4 + C1 + C2 + C3)\end{tabular} & 0.4281            \\ \hline
\begin{tabular}[c]{@{}c@{}}Chakma et al.~\cite{3rd}\\ (2023)\end{tabular} & \begin{tabular}[c]{@{}c@{}}BanglaBERT-Large with \\ Random Token Drop\end{tabular}           & 0.7330            \\ \hline
\begin{tabular}[c]{@{}c@{}}Kabir et al.~\cite{2nd}\\ (2024)\end{tabular}  & BanglaBERT (large)                                                                           & 0.8273            \\ \hline
\end{tabular}%
}
\end{table}

When comparing our best-performing model, the Decision Tree with AdaBoost (F1-score of 0.7860), with other studies using the EmoNoBa dataset, it outperforms Islam et al. (2022)~\cite{8}, who achieved an F1-score of 0.4281 using a lexical feature model. However, it falls short of the results from Kabir et al. (2024)~\cite{3rd}, who achieved an F1-score of 0.8273 with BanglaBERT (large), and Chakma et al. (2023)~\cite{2nd}, who achieved a 0.7330 F1-score using BanglaBERT-Large with Random Token Drop. This indicates that while our model performs well, state-of-the-art models like BanglaBERT offer superior results on this dataset.

\section{Limitations and Future Works}
Despite being insightful, our study had certain drawbacks. Only social media was used to collect the data, which does not adequately represent how Bangla is used in everyday life or in various contexts. Additionally, certain emotions were more prevalent than others in the data, which might have had an impact on our findings. Furthermore, the text analysis tools we employed, such as n-grams, may not have fully captured all the subtleties of the language.
Our goal is to obtain a more comprehensive understanding of Bangla usage in the future by utilizing a greater range of texts. In order to fairly represent all emotions, we also want to balance the data. We intend to test more recent technologies, such as BERT or GPT, which may be more adept at comprehending context. Enhancing our pre-analysis text processing techniques may also be beneficial. Investigating alternative strategies for fusing various models could provide us with a better understanding of Bangla and emotions.

\section{Conclusion }
The study took important steps in understanding emotions in Bangla text through social media comments. We used several machine learning models and found that the Decision Tree performed the best, with and without the AdaBoost the performance of it was almost identical. Our study adds to the field of sentiment analysis in Bangla and points to promising directions for future research to improve accuracy and understanding of emotional expressions in language.

\end{document}